\documentclass[sigconf,screen=true]{acmart}

\renewcommand\footnotetextcopyrightpermission[1]{}
\setcopyright{none}
\usepackage{balance}
\usepackage{courier}
\usepackage{empheq}
\usepackage{xcolor}
\usepackage{amsmath,amsfonts}
\usepackage{graphicx}
\usepackage{enumitem}
\usepackage{textcomp}
\usepackage{multirow}
\usepackage{colortbl}
\usepackage{booktabs}
\usepackage{bm}
\usepackage{siunitx}
\usepackage{microtype}
\usepackage{flushend, epstopdf}
\usepackage{adjustbox}
\usepackage{mathtools,algorithm}
\usepackage{algpseudocode}

\usepackage{arydshln}
\usepackage[skip=1pt]{caption}
\usepackage{subcaption}
\DeclareSIUnit\GE{GE}
\makeatother
\newcommand{\n}{\mathcal{N}}

\newcolumntype{C}[1]{>{\centering\let\newline\\\arraybackslash\hspace{0pt}}m{#1}}
\newcolumntype{H}{>{\setbox0=\hbox\bgroup}c<{\egroup}@{}}
\newcommand{\ubf}[1]{\underline{\textbf{#1}}}

\definecolor{NVgreen}{RGB}{118,185,0}
\definecolor{NVblack}{RGB}{0,0,0}
\definecolor{NVlgrey}{RGB}{205,205,205}
\definecolor{NVmgrey}{RGB}{140,140,140}
\definecolor{NVdgrey}{RGB}{94,94,94}

\definecolor{NVemerald}{RGB}{0,133,100}
\definecolor{NVamethyst}{RGB}{93,22,130}
\definecolor{NVintel}{RGB}{0,113,197}
\definecolor{NVgarnet}{RGB}{137,12,88}
\definecolor{NVfluorite}{RGB}{250,194,0}

\definecolor{cardinal}{rgb}{0.77, 0.12, 0.23}
\definecolor{royalazure}{rgb}{0.0, 0.22, 0.66}
\definecolor{goldenyellow}{rgb}{1.0, 0.87, 0.0}
\graphicspath{{./fig/}}

\AtBeginDocument{%
  }

\begin{document}

\title{GOALPlace: Begin with the End in Mind}

\author{Anthony Agnesina}
\affiliation{\institution{NVIDIA}\country{}}
\author{Rongjian Liang}
\affiliation{\institution{NVIDIA}\country{}}
\author{Geraldo Pradipta}
\affiliation{\institution{NVIDIA}\country{}}
\author{Anand Rajaram}
\affiliation{\institution{NVIDIA}\country{}}
\author{Haoxing Ren}
\affiliation{\institution{NVIDIA}\country{}}
\renewcommand{\shortauthors}{Agnesina et al.}

\begin{abstract}
Co-optimizing placement with congestion is integral to achieving high-quality designs. This paper presents GOALPlace, a new learning-based general approach to improving placement congestion by controlling cell density. Our method efficiently learns from an EDA tool's post-route optimized results and uses an empirical Bayes technique to adapt this goal/target to a specific placer's solutions, effectively beginning with the end in mind. It enhances correlation with the long-running heuristics of the tool's router and timing-opt engine---while solving placement globally without expensive incremental congestion estimation and mitigation methods. A statistical analysis with a new hierarchical netlist clustering establishes the importance of density and the potential for an adequate cell density target across placements. Our experiments show that our method, integrated as a demonstration inside an academic GPU-accelerated global placer, consistently produces macro and standard cell placements of superior or comparable quality to commercial tools. Our empirical Bayes methodology also allows a substantial quality improvement over state-of-the-art academic mixed-size placers, achieving up to 10$\times$ fewer design rule check (DRC) violations, a 5\% decrease in wirelength, and a 30\% and 60\% reduction in worst and total negative slack (WNS/TNS).
\end{abstract}

% \begin{CCSXML}
% <ccs2012>
% <concept>
% <concept_id>10010583.10010682.10010697</concept_id>
% <concept_desc>Hardware~Physical design (EDA)</concept_desc>
% <concept_significance>500</concept_significance>
% </concept>
% <concept>
% <concept_id>10010583.10010682.10010697.10010701</concept_id>
% <concept_desc>Hardware~Placement</concept_desc>
% <concept_significance>500</concept_significance>
% </concept>
% </ccs2012>
% \end{CCSXML}

% \ccsdesc[500]{Hardware~Physical design (EDA)}
% \ccsdesc[500]{Hardware~Placement}


\maketitle
\section{Introduction}\label{sec1}

\begin{figure}[t]
\centering
\includegraphics[width=\columnwidth]{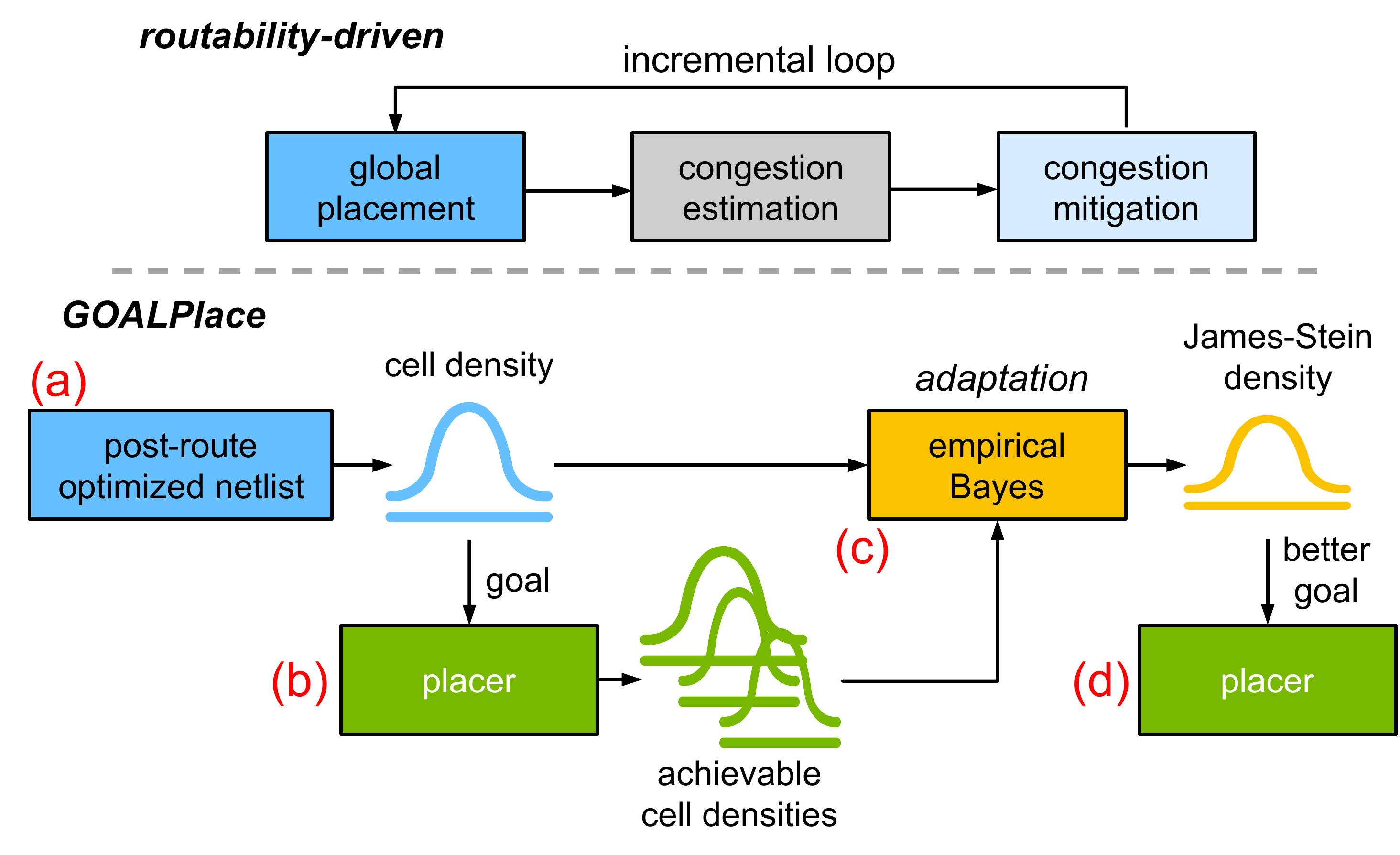}
\caption{The traditional incremental routability-driven method vs. GOALPlace, our placement congestion mitigation approach based on cell density and empirical Bayes. Key steps include: (a) Generate a learning goal from a post-route optimized netlist; (b) Use the specific placer to generate cell densities based on the goal from Step (a); (c) Apply the empirical Bayes approach to refine the goal for the placer; (d) Utilize this refined goal to produce high-quality placements. This methodology is versatile and applicable to any tool flow and placer.}
\Description[]{}
\label{fig:flows}
\end{figure}

Placement significantly impacts congestion/routability in modern designs and technology nodes due to the ever-increasing cell density, design rules complexity, and amount of macros~\cite{alpert2010makes}. Routability-driven placement is a well-developed field where most methods follow the two-step incremental approach~\cite{markov2012progress} (often integrated within a loop, as illustrated at the top of Figure~\ref{fig:flows}); congestion is estimated first through various methods and then mitigated by cell density control. However, incremental methods struggle to adjust the overall placement distribution towards an ideal wirelength and congestion trade-off, especially for mixed-size placement where local density changes can unexpectedly disrupt the global landscape. More importantly, their success heavily relies on the congestion estimation's accuracy, which ultimately depends on a strong correlation with the router's long-running heuristics. This limitation currently impedes the use of predictive models in industrial settings to replace costly but accurate algorithms.

Cell density is the most notable change brought by routability and timing optimizations. It is critical because it determines the placement congestion~\cite{viswanathan2010itop}, i.e., where cells and macros are heavily packed in local regions. The effects of placement congestion are two-fold: (1) it restricts available space for the optimization step, such as gate sizing and buffer insertion, thereby degrading its performance, and (2) it can cause severe routing congestion. Industrial methodologies often resort to clever placement ``conditioning'' with density constraints like screens and blockages to improve quality metrics~\cite{kahng2024solvers}. However, robust methods for manually synthesizing such ``magic'' constraints are lacking. We believe learning-based methods can be a powerful tool for efficiently generating cell density constraints/goals/targets for global placement.

In this work, we propose \textit{GOALPlace}, a general methodology that learns from post-route optimized cell density, capturing essential information about the high-quality optimizations performed across the flow. Optimizing directly at the placement stage for the right density, i.e., \textit{beginning with the end in mind}, improves the correlation with the tool's assumptions and final results. Our methodology is very data efficient and adaptable to any tool flow and placer, commercial and academic. We achieve the desired distribution by leveraging the cell density targets to inflate cells during global placement, eliminating the need for congestion estimation and incremental behavior. We \textit{ground} the density targets in reality by calibrating them to the achievable outputs of the selected placer using an empirical Bayes method~\cite{efron2012large}. Figure~\ref{fig:flows} summarizes our GOALPlace flow compared to the incremental routability-driven method. The key contributions are as follows:
\begin{itemize}
\item We propose controlling the cell density during mixed-size global placement to improve placement congestion. Our prototypical method does not require congestion estimation and escapes the limitations of incremental approaches. 
\item We efficiently learn each design's cell density distribution, relying on a unique previous run of a commercial EDA tool. This improves correlation with its routing and timing optimization engines.
\item We introduce an empirical Bayes method to adjust the initial cell density targets derived from the tool's post-route results to achieve improved distributions tailored for the specific placer. Furthermore, we propose a theoretical refinement to the Bayes estimator, specifically targeting timing considerations.
\item We conduct a statistical analysis using a new hierarchical netlist clustering to demonstrate a strong correlation between placement, density, and timing, underscoring the importance of density. The consistent clustering effect and stable densities suggest generic cell density targets across placements in a fixed design context.
\item We integrate GOALPlace inside DREAMPlace~\cite{lin2019dreamplace} and the AutoDMP methodology~\cite{agnesina2023autodmp} to demonstrate our approach. Experimental results on academic and industrial flows and benchmarks show that our solution consistently produces macro and standard cell placements of superior or comparable post-route quality to the commercial tool.  Furthermore, we achieve up to 10$\times$ fewer design rule check (DRC) violations, a 5\% decrease in wirelength, and a 30\% and 60\% reduction in worst and total negative slack (WNS/TNS) compared to the default DREAMPlace/AutoDMP baseline.
\end{itemize}

\section{Background and Motivations}\label{sec2}

This section outlines advanced methods for improving placement congestion and motivates the importance of density with a statistical analysis of placements using a new hierarchical netlist clustering.

\subsection{Related Work}

Current routability-driven placement methods range from traditional incremental approaches to newer, learning-based schemes. Each has notable shortcomings, which we detail below.

\subsubsection{Congestion Estimation and Mitigation}\label{subsec:cong-est}
Accurate measurement of congestion is crucial to the success of routability-driven placement, given that it dictates actionable measures for mitigation. Most estimation methods seek to build accurate congestion maps to identify complex routing regions. The most accurate but expensive approach is constructive, which invokes high-quality global routers~\cite{roy2009crisp,brenner2015bonnplace}. While advances in fast global routers are emerging, different routers will generate different congestion maps, leading to divergent quality of results (QoR) when the placements are used with a different router. A second approach routes net topologies with a probabilistic model such as RUDY~\cite{spindler2007fast}, but is less popular due to its inaccuracy in the growing complexities of modern designs. 

Once an initial placement is analyzed for congestion, it is gradually refined. A common idea in global placement is to inflate cells in over-congested areas to reduce placement density locally~\cite{brenner2002effective}. The cell inflation factor is relative to how congested the cell location is, sometimes also accounting for pin density. Additionally, some methods integrate congestion data directly in their objective function~\cite{liu2021global}. Manual density control methods, e.g., density screens, selective module-level or cell-level padding, and global uniform density targets, are highly inflexible, and predicting their unwanted consequences on timing QoR is challenging. Additionally, all the methods mentioned are less effective at mitigating congestion without pre-fixed macros and can deviate designs from optimal wirelength when applied incrementally. They also struggle to remove congestion on high-density designs with limited free space, where density is better optimized globally.

\subsubsection{Newer Learning-Based Approaches}
Predictive AI methods can replace expensive calls to the global router in congestion estimation by predicting congestion counts and hotspots~\cite{yan2022towards}. However, these methods are not robust to noise and require large amounts of high-quality commercial data, making them less applicable to industrial settings and diverse test cases.

The method in~\cite{lu2023dream} employs generative adversarial learning to generate a density map during placement that mimics the ones of commercial tools. However, its learning target lacks clarity. It is obtained from the placement stage and aims to replicate the density map globally without considering individual cells. This approach also relies on a substantial database of prior designs. Another work~\cite{cai2023puffer} predicts cell inflation based on features from congestion maps but depends on many heuristic techniques for map creation and cell padding. In contrast, our method, presented in detail later, is grounded in physical reality and targeted, aiming to match an EDA tool's final cell density results at the global placement stage. It also readily integrates into any tool flow and placer using cell-based inflation without algorithmic modifications.

\subsection{Density Analysis with Clustering}\label{subsec:cluster}

\begin{table}[t]
\centering\small
\setlength{\tabcolsep}{1.5pt}
\caption{Clustering analysis on four TILOS~\cite{cheng2023assessment} benchmarks reveals correlations between cluster density and timing criticality. $\rho_{DT}$ represents the average correlation, while  $\rho_{DT,\mathrm{tim-crit.}}$ is restricted to the most timing-critical clusters. Notably, assuming a Gaussian distribution, the density of a cluster across placements falls within its mean $\pm0.85\sigma$ with a probability exceeding 80\%.}\label{tab:cluster-corr}
\begin{adjustbox}{width=\columnwidth}
\begin{tabular}{|c||c|c|c|c|c|c|c|}
\hline
\rowcolor{NVlgrey}
\textbf{Design}&\textbf{\#cells}&\textbf{\#clusters}&$\boldsymbol{\rho_{DT}}$&$\boldsymbol{\rho_{DT,\mathrm{tim-crit.}}}$&$\boldsymbol{\sigma_{\mathrm{cluster-dens.}}}$&\textbf{DBI}&\textbf{RT}\\
\hline\hline
Ariane&100K&8&0.69&0.84&0.08&1.55&11s\\
NVDLA&150K&10&0.77&0.88&0.11&1.21&16s\\
BlackParrot&650K&25&0.61&0.71&0.09&2.05&22s\\
MemPool Group&2.7M&69&0.72&0.91&0.10&2.40&31s\\
\hline
\end{tabular}
\end{adjustbox}
\end{table}

\begin{figure}[t]
\centering
\includegraphics[width=\columnwidth]{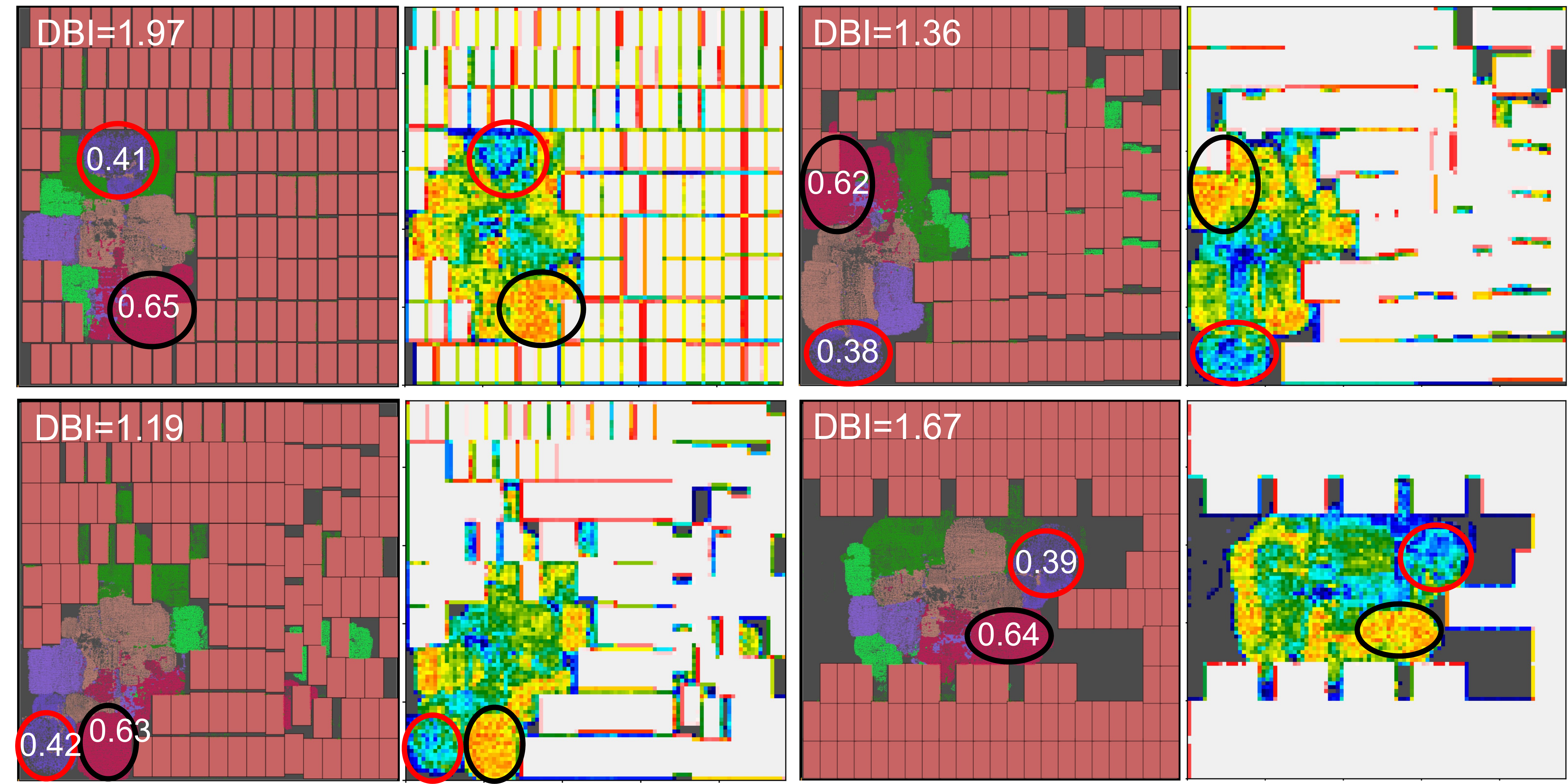}
\caption{Visual analysis of our netlist clustering on Ariane NanGate45 benchmark, with the DBI values indicated. Densities and proximity of clusters are maintained despite their rearrangement.}
\Description[]{}
\label{fig:density}
\end{figure}

The authors of~\cite{fogacca2019finding} demonstrated the ability to predict netlist clusters that ``stay together'' throughout implementation, which is crucial for understanding the general aspects and impacts of density on metrics like timing. Recognizing the complexity of factors affecting the fine-grained cell-level post-route density, the cluster's view allows for a broader, coarse-scale perspective, which we find more convincing and generalizable. Nevertheless, the method in~\cite{fogacca2019finding} fails to generate clusters with apparent density and timing correlations, likely due to algorithmic deficiencies. Our proposed clustering, which includes hierarchy information, aims to establish these correlations more effectively, focusing solely on revealing statistical properties. Please note that the clustering is provided for motivation and analysis only and will not be used in our algorithm, which operates at the cell level.

\subsubsection{Clustering Method}
We propose clustering at the whole module/unit level to prevent merging cells without a common hierarchical parent, unlike the flat netlist approach in~\cite{fogacca2019finding}. Our method respects RTL's logical groups and timing boundaries, using a top-down traversal to identify modules for clustering, stopping when their instance counts fall within a predetermined range. Instances lacking hierarchy are bundled together for clustering. This clustering is practical in industrial ASIC design where the netlist logical hierarchy is commonly preserved during physical implementation to simplify DFT and verification.

We select the Leiden algorithm~\cite{traag2019louvain} over Louvain~\cite{fogacca2019finding} for module decomposition because it is faster and effectively resolves issues of poorly connected communities found by Louvain. We apply clustering to the clique-expansion graph of the hypergraph netlist, with parameters like Leiden's resolution, RTL module size to break down, and edge weight, tuned to balance cluster count, the Davies–Bouldin index (DBI)~\cite{davies1979cluster} measuring physical closeness within a cluster, and the correlation between cluster density (=average of cell densities in a cluster) and timing criticality (=average of cell slacks). Overall, the Tsay-Kuh clique weights~\cite{tsay1991unified} complemented with a normalized Levenshtein string distance~\cite{yujian2007normalized} is the most effective:
\begin{equation}\label{eq:clique}
d(\{u,v\}\in e)=\dfrac{2}{|e|\cdot\big(1+\operatorname{Levenshtein}(u, v)\big)},
\end{equation}
where the Levenshtein distance emphasizes hierarchical attraction. Our implementation includes a fast clique-expansion graph method using sparse matrix multiplication~\cite{agarwal2006higher} on GPUs, a hierarchical trie for RTL traversal, and cuGraph's GPU-based Leiden algorithm~\cite{fender2022rapids}. Table~\ref{tab:cluster-corr} shows it scales very well, clustering netlists up to 3M nodes/36M clique-edges (MemPool Group) in under 30s on an NVIDIA A100.

\subsubsection{Density Analysis}
We conduct a statistical analysis of clustering on four benchmarks detailed in Table~\ref{tab:cluster-corr}. We analyze four distinct post-route optimized results from a commercial EDA tool for each benchmark, with diverse global placements originating from DREAMPlace and the tool. Our clustering accurately identifies communities of standard cells that consistently co-locate across placements, evidenced by low DBI values and visualized in Figure~\ref{fig:density}, confirming that cells remain together despite rearrangements. We calculate each cluster's average cell density and timing criticality, uncovering a strong correlation ($\rho_{DT}$) between lower density and higher timing criticality, particularly in critical clusters ($\rho_{DT,\mathrm{tim-crit.}}$). This supports our strategy of maintaining strict low density for timing-critical cells in our empirical Bayes approach. Additionally, minor standard deviations ($\sigma_{\textrm{cluster-dens.}}$) indicate consistent cluster-level densities, as shown in Figure~\ref{fig:density}, which depicts minimal fluctuations in two highlighted clusters. Given the influence of nearby cells on a cell's density, this consistency supports the feasibility of adequate cell-level density targets for a given design and context to enhance placement outcomes. Our experiments will show that although our approach concentrates on local cell-level control, it also influences density at the broader cluster level due to the placer's clustering effect.

In our methodology, we prefer cell-level density constraints over cluster-level for two key reasons: (1) Cell-level noise is averaged across many cells, minimizing its effects, whereas cluster-level noise does not amortize, causing significant issues. Furthermore, a few suboptimal cell-level targets can be fixed easily by the EDA tool's incremental optimizer; (2) Our empirical Bayes approach provides significant benefits in large-scale settings ($\#\mathrm{cells}\gg\#\mathrm{clusters}$).

\section{Methodology}\label{sec3}

This section presents GOALPlace's key components, illustrated in Figure~\ref{fig:flows}. First, we explain the derivation of cell density targets from post-route placement data of an EDA tool. Next, we apply an empirical Bayes method to adjust these targets to densities achievable by a specific placer. Finally, we illustrate the application of our methodology through cell inflation-based density control in the academic global placer DREAMPlace, integrated with the new ML-based AutoDMP search. 

Due to its broad applicability, simplicity, and speed, GOALPlace can be effectively integrated into any tool flow and placer that supports cell inflation for density control. Additionally, it is more data-efficient than traditional model-based machine learning methods, requiring only one post-route optimized result from an EDA tool and a few runs of the targeted placer for a given design.

\subsection{Target Density Selection}

\begin{figure}[t]
\centering 
\includegraphics[width=.8\columnwidth]{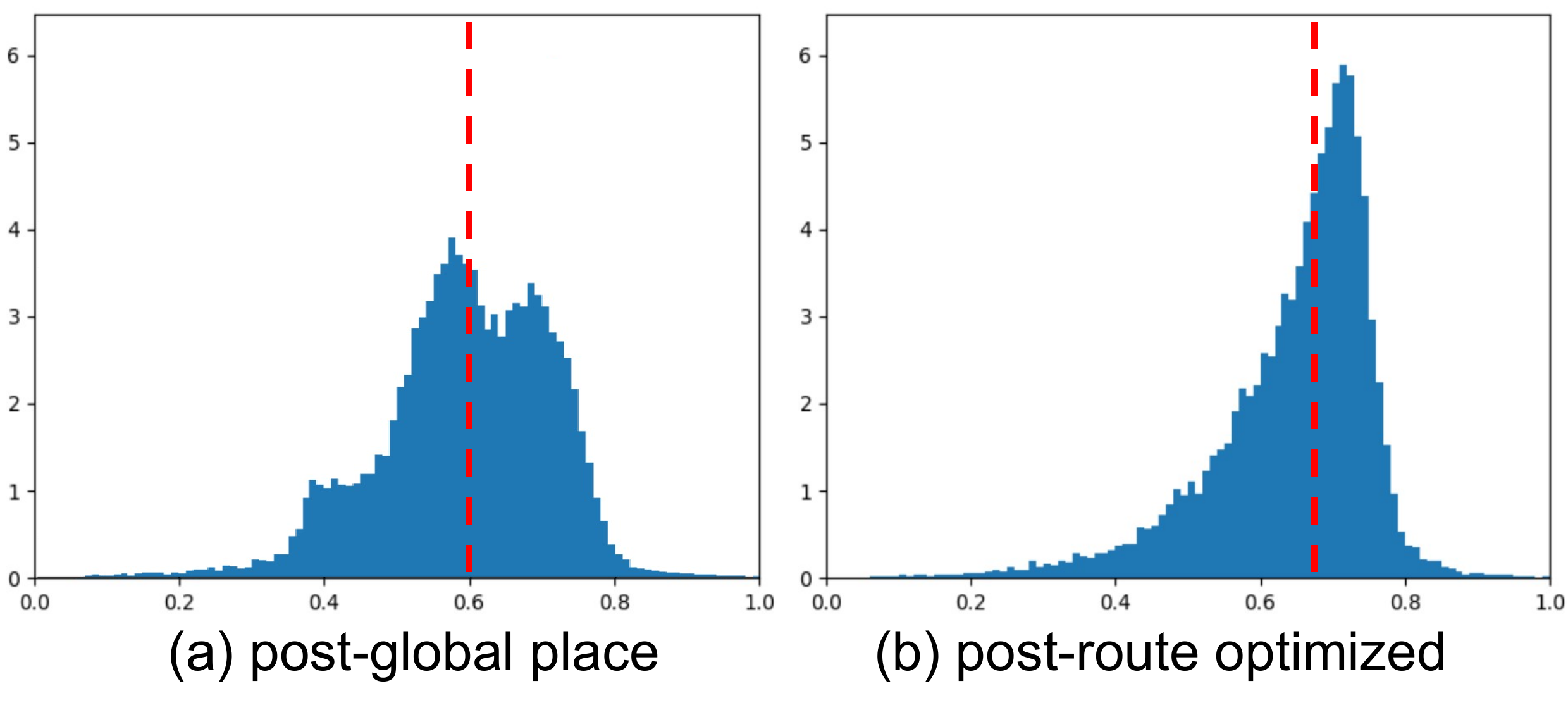}
\caption{Noticeable shift in cell density between global and post-route stages on the BlackParrot NanGate45 benchmark, with the post-route density providing more valuable information into later-stage timing and congestion optimizations.}
\Description[]{}
\label{fig:shift}
\end{figure}

Figure~\ref{fig:shift} illustrates density changes between global placement and post-route optimization, considering only cells present in both stages and resized to post-synthesis sizes. The density shift thus results from placement adjustments rather than adding new cells or gate resizing. The rise in average cell density, driven by the timing-driven placement that pulls cells together, suggests that post-route cell density--reflecting later-stage timing and congestion optimizations--contains more vital information and should be our target.

In our experiments, we rely on a single run of an EDA tool to post-route optimization to establish the target cell densities. Such data is readily available in industrial flows, especially with the emergence of design space exploration tools like Synopsys DSO.ai~\cite{verma2024dso} and Cadence Cerebrus~\cite{tan20241}. Notably, the global placement pushed to post-route to generate the cell density targets can originate from any commercial or academic placer. We maintain the targets fixed across all AutoDMP runs, even with movable macros. This approach is justified by findings outlined in Section~\ref{subsec:cluster} and the assumption of a consistent design context, i.e., identical floorplan shape, IO placement, routing layers, and timing constraints.

\subsection{Target Density Computation}
To compute the cell density, we discretize the floorplan into a 2D grid of bins $B$. For any bin $b\in B$ and cell $i\in V$, let $\operatorname{OA}(i,b)$ be their overlap area, and $A_b$ and $a_i$ their respective area. The bin and cell density are
\begin{equation}\label{eq:bin-density}
\rho_b=\sum_{i\in V}\frac{\operatorname{OA}(i, b)}{A_b}\quad\mbox{and}\quad\rho_i=\sum_{b \in B}\rho_b\frac{\operatorname{OA}(i, b)}{a_i}.
\end{equation}
In all our experiments, we compute the cell density using square bins of size 10$\times$10 the standard cell row height.

We compute the density targets of the cells in the place netlist from post-route optimized results (Step (a) in Figure~\ref{fig:flows}) as follows: (1) Buffers are removed; (2) Cells in both place and post-route netlists are placed in their post-route positions (including macros) and sized to their post-synthesis size; (3) The remaining standard cells that do not match are set to``zero'' size---in practice, site width and row height. By setting these to zero size, we provision space for their replacement with other cells found in the post-route netlist, effectively decreasing the target density of matching cells in the place netlist. This is particularly important to account for the space needed by later timing-opt steps that insert buffers and upsize and pull together critical cells. Moreover, timing-driven placement engines are also known to pad cells on critical paths. The non-matching cells are also set to zero size during global placement and for density map comparison in AutoDMP. They are, however, reverted to the correct size for legalization and detailed placement.

\subsection{Target Density Adaptation with Empirical Bayes}

\begin{figure}[t]
\centering
\begin{subfigure}[b]{0.45\columnwidth}
\centering
\includegraphics[trim={0 10 30 50},clip,width=\textwidth]{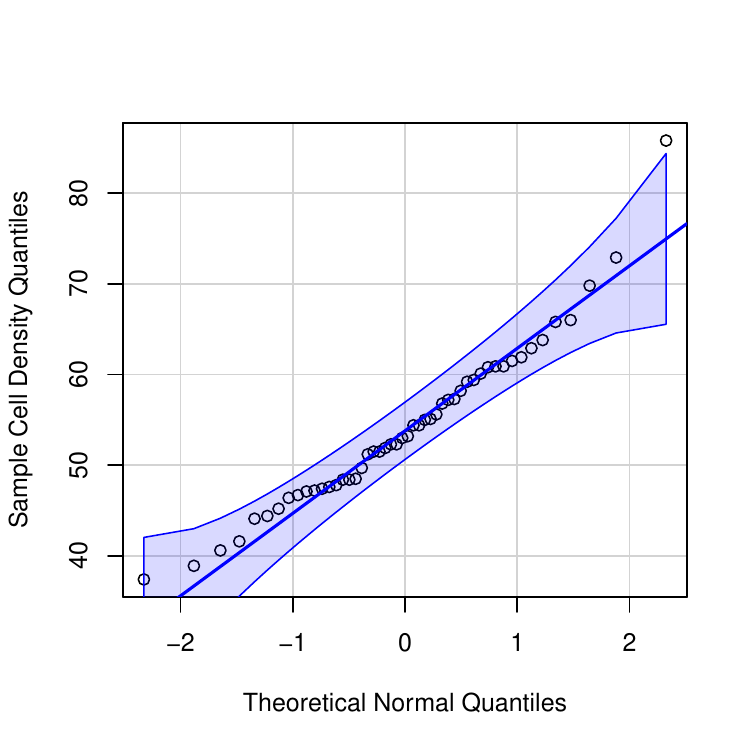}
\caption{}
\label{fig:quantile}
\end{subfigure}
\hfill
\begin{subfigure}[b]{0.54\columnwidth}
\centering
\includegraphics[trim={0 5 5 5},clip,width=\textwidth]{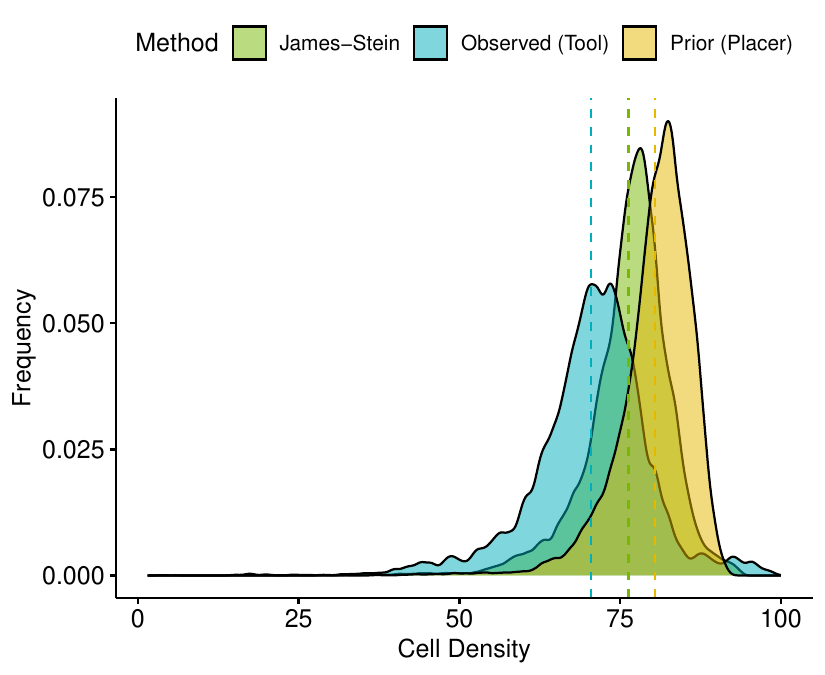}
\caption{}
\label{fig:JS}
\end{subfigure}
\caption{(a) The Quantile-Quantile plot of the density of one cell across the Pareto-optimal placements follows a Gaussian distribution. (b) Using the empirical Bayes James-Stein estimator on the Ariane ASAP7 benchmark to learn optimal cell density targets $\boldsymbol{\hat\mu^{(\mathrm{JS})}}$ (green) by shrinking the tool distribution $\boldsymbol{z}$ (blue) towards achievable prior densities of DREAMPlace $\boldsymbol{\hat\mu^{(\mathrm{P})}}$ (yellow).}
\Description[]{}
\label{fig:quantile-JS}
\end{figure}

While we will use cell inflation as guidance to enforce the target densities, it is important to note that the selected placer may struggle to achieve post-route densities from the EDA tool while optimizing for other objectives like wirelength. This limitation may arise from integrating different algorithms and the complexity of replicating post-route densities, which result from multiple discrete optimizations and various factors. We also found that overly strict density control methods, more intrusive than inflation, can substantially degrade wirelength. Hence, we propose adapting targets to a specific placer by learning from its placement solutions rather than imposing a possibly unrealistic density distribution.

\subsubsection{Placer Density Distribution}
To determine the achievable densities for our placer, DREAMPlace, we will use the top 50 Pareto global placements and their achieved cell densities from an AutoDMP search using inflation with the original tool's targets (Step (b) in Figure~\ref{fig:flows}). We focus on Pareto points to ensure diversity and high quality in the placements. Note that we must first bias DREAMPlace towards better density distributions through inflation and learn from these outcomes, as default DREAMPlace’s uniform-density solutions offer limited insights.

\subsubsection{Mathematical Framework}
Let $z_{ik}$ denote the density of cell $i$ from the $k$-th Pareto placement (here $k=1,2,\dots,50$). Interestingly, for each cell, the cell density across placements closely follows a Gaussian distribution with small variance; on all benchmarks, more than 50-70\% of the cells have their density distributed as in the Quantile-Quantile plot in Figure~\ref{fig:quantile}. This behavior is important to the following theoretical approach.

Suppose each cell density has a true expectation $\mu_i$ from ``nature'', which we wish to estimate having observed the tool density $z_i$ and considering the placer's placements prior. The prior encodes our preconceived belief that the target should behave like the placer's densities. We assume normally distributed cell densities according to the previous observation,
\begin{equation}\label{eq:gaussian-model}
z_i\sim\n(\mu_i,\sigma_0^2)\quad(i=1,2,\dots,N),
\end{equation}
where $N$ is the number of cells, and taking variance $\sigma_0^2$ as known (e.g., $=\text{var}\{z_i\}$). Without any prior knowledge of how differently the unknown parameters $\mu_i$ might arrange themselves, the obvious estimators are
\begin{equation}
\hat\mu_i^{(\mathrm{MLE})}=z_i\quad\mbox{for}\,\,i=1,2,\dots,N,
\end{equation}
the ``MLE'' noting that these are maximum likelihood estimates. 

\subsubsection{James--Stein Estimator}
This is where the James--Stein estimator~\cite{efron1973stein}, an empirical Bayes technique, enters (Step (c) in Figure~\ref{fig:flows}). It envisioned a ``normal-normal'' version,
\begin{equation}\label{eq:model}
\mu_i\sim\n(M_i,A)\quad\mbox{and}\quad z_i|\mu_i\sim\n(\mu_i,\sigma_0^2)
\end{equation}
where both $M_i$ and $A$ are unknown. Now, the posterior expectation of $\mu_i$ given $z_i$, which is the Bayes estimator under squared error loss of model \eqref{eq:model} is 
\begin{equation}\label{eq:bayes}
\hat\mu_i^{(\mathrm{Bayes})}=M_i+B(z_i-M_i)\quad\quad\big[B=A/(A+\sigma_0^2)\big].
\end{equation}
The James--Stein empirically estimates parameters $M_i$ and $B$ in \eqref{eq:bayes} unbiasedly from targets $z_i$ and prior knowledge $z_{ik}$,
\begin{equation}\label{eq:bhat}
\widehat{M}_i=\hat\mu_i^{(\mathrm{P})}=\sum_{k=1}^{50} z_{ik}\big/50\quad\mbox{and}\quad \widehat{B}=\Bigg[1-\frac{(N-2)\sigma_0^2}{S}\Bigg],
\end{equation}
where $S=\sum \big(z_i-\hat\mu_i^{(\mathrm{P})}\big)^2$. The James--Stein rule follows
\begin{equation}\label{eq:JS}
\hat\mu_i^{(\mathrm{JS})}=\hat\mu_i^{(\mathrm{P})}+\underbrace{\biggl(1-\frac{(N-2)\sigma_0^2}{S}\biggr)}_\textrm{shrinking factor} \Big(z_i-\hat\mu_i^{(\mathrm{P})}\Big).
\end{equation}
Figure~\ref{fig:JS} shows how the James--Stein embodies a ``middle-ground'' between the prior and target, by shifting the target toward the placer's densities. If the shrinking factor were equal to 1, then the James--Stein estimator for a given cell is identical to that cell's tool density. Stein's theorem states that the shrinking factor is always less than 1. So when the target density of a cell is much higher than what the placer can achieve, it should be reduced, and vice versa. This makes unachievable targets more reasonable. Importantly, one can show the following striking theorem proved by James and Stein:
\begin{theorem}\label{thm1}
For $N\geq3$, the James--Stein estimator everywhere dominates the MLE in terms of expected total squared error (the ``risk'' $R$); that is,
\begin{equation}
E_{\textrm{$\boldsymbol{\mu}$}}\Big\{\Vert\textrm{$\boldsymbol{\hat\mu}$}^{(\mathrm{JS})}-\textrm{$\boldsymbol{\mu}$}\Vert^2\Big\} < E_{\textrm{$\boldsymbol{\mu}$}}\Big\{\Vert\textrm{$\boldsymbol{\hat\mu}$}^{(\mathrm{MLE})}-\textrm{$\boldsymbol{\mu}$}\Vert^2\Big\}
\end{equation}
for \emph{every} choice of $\textrm{$\boldsymbol{\mu}$}$.
\end{theorem}
This theorem can be proved from Stein's unbiased risk estimate~\cite{tibshirani2015stein}, by which we also compare the James--Stein risk to the Bayes risk
\begin{equation}\label{eq:risks}
R^{(\mathrm{JS})}\Big/ R^{(\mathrm{Bayes})}=1+\frac{2\sigma_0^2}{NA}.
\end{equation}
Result \eqref{eq:risks} confirms our motivations for cell level matching---it implies that even if the signal-to-noise ratio is low ($A\ll\sigma_0^2$), the James--Stein risk converges quickly towards optimality as the number of cells $N$ increases.

The improvement of this empirical Bayes technique resides in its large-scale inference by ``learning from the experience of others''. For each cell, the James--Stein estimator pools together the information of all the other cells in the design, making it extremely data efficient (i.e., a single post-route result is assumably enough). The experiments will show that these new densities significantly enhance placement congestion and wirelength when used to inflate cells inside the placer (Step (d) in Figure~\ref{fig:flows}). However, we do observe timing degradation in some cases, which we aim to solve with the following enhancement.

\subsubsection{Enhancement for Timing}\label{subsec:timing}
The James--Stein theorem concentrates attention on the total squared error loss function $\sum (\hat\mu_i-\mu_i)^2$, without concern for the effects on individual cells. While most of those effects are good, genuinely unusual cases, like low-density timing-critical cells, can suffer and should not have been shrunk so drastically toward the mean. We propose a compromise method that captures most of the group savings while protecting unusual individual cases of timing-critical cells with a more targeted inflation. We decide to follow the James--Stein estimate \eqref{eq:JS} subject to the restriction of not deviating more that $D_i\sigma_0$ units away from $\hat\mu_i^{(\mathrm{MLE})}=z_i$, where $D_i$ is set according to the timing-criticality of the cell;
\begin{equation}\label{eq:translation}
\hat\mu_i^{(\mathrm{JS/D})}=\begin{cases}
\max\Big(\hat{\mu}_i^{(\mathrm{JS})},\hat\mu_i^{(\mathrm{MLE})}-D_i\sigma_0\Big)&\text{for $z_i>\hat\mu_i^{(\mathrm{P})}$}\\
\min\Big(\hat{\mu}_i^{(\mathrm{JS})},\hat\mu_i^{(\mathrm{MLE})}+D_i\sigma_0\Big)&\text{for $z_i\le\hat\mu_i^{(\mathrm{P})}$}.
\end{cases}
\end{equation}
In practice, we propose to set $D_i$ as the 10-quantile post-route slack value of each cell. For the most timing-critical cells, $D_i=1$, the smallest, which says that $\hat\mu_i^{(\mathrm{JS/D})}$ will never deviate more than $\sigma_0$ from $z_i$, putting a higher tightness to respect the density target. It will be shown in the experiments that using $\boldsymbol{\hat\mu}^{(\mathrm{JS/D})}$ instead of $\boldsymbol{\hat\mu}^{(\mathrm{JS})}$ to inflate the cells inside the placer (Step (d) in Figure~\ref{fig:flows}) considerably improves post-route timing.

\subsubsection{Extension to Different Variances}
For mathematical curiosity, we briefly discuss a more realistic but complex situation than in \eqref{eq:gaussian-model},
\begin{equation}
z_i\sim\n(\mu_i,\sigma_i^2)\quad(i=1,2,\dots,N),
\end{equation}
where the $\sigma_i$ are known, but are different from one another. In practice, most cells show small and similar density variances across placements ($<10\%$), but some exhibit much larger variations. We continue to assume $\mu_i\sim\n(M_i,A)$, which leads to 
\begin{equation}\label{eq:bayes2}
\hat\mu_i^{(\mathrm{Bayes})}=M_i+B_i(z_i-M_i)\quad\quad\big[B_i=A/(A+\sigma_i^2)\big].
\end{equation}
An equation for the James--Stein estimator in this setting has yet to be found; however, the authors in~\cite{efron1973stein} developed a method to calculate the shrinking factor which we present in Algorithm~\ref{alg:shrinking} for the sake of completion.

\begin{algorithm}
\caption{Shrinking Factor for Different Variances}\label{alg:shrinking}
\begin{algorithmic}[1]
\Procedure{Shrinking Factor}{cell i}
    \State (1) Set $d_i=3$, $d_j=1$, $j\neq i$.
    \State (2) Then $E_i=(S_i-d_i\sigma_i^2)/d_i$, w/ $S_i=\big(z_i-\hat\mu_i^{(\mathrm{P})}\big)^2$, is an unbiased estimator of $A$ with variance $1/I_i(A)$, $I_i(A)$ being the Fisher information for $A$ in $S_i$: $I_i(A)=d_i/2(A+\sigma_i^2)^2$.
    \State (3) Solve for $\hat{A}=(\sum_j E_j\cdot I_j(\hat{A}))/\sum_j I_j(\hat{A})$.
    \State (4) Define $d_i^*=2(\hat{A}+\sigma_i^2)^2\sum_j \frac{d_j}{2(\hat{A}+\sigma_j^2)^2}$.
    \State \textbf{return} $\hat{B}_i=1-\frac{d_i^*-4}{d_i^*}\frac{\sigma_i^2}{\hat{A}_i+\sigma_i^2}$ .
\EndProcedure
\end{algorithmic}
\end{algorithm}

In the cases where all the $\sigma_i$ equal $\sigma_0$, we have $\hat{A}=S/(N+2)$ and $d_i^*=(N+2)$, which gives $\hat{B}$ coinciding exactly with \eqref{eq:bhat}.

\subsection{Cell Density Enforcement}

After establishing a cell density target, we now present how to enforce it using cell inflation inside our selected placer DREAMPlace.

\subsubsection{ePlace Uniform Density}

\begin{figure}[t]
\centering 
\includegraphics[width=.8\columnwidth]{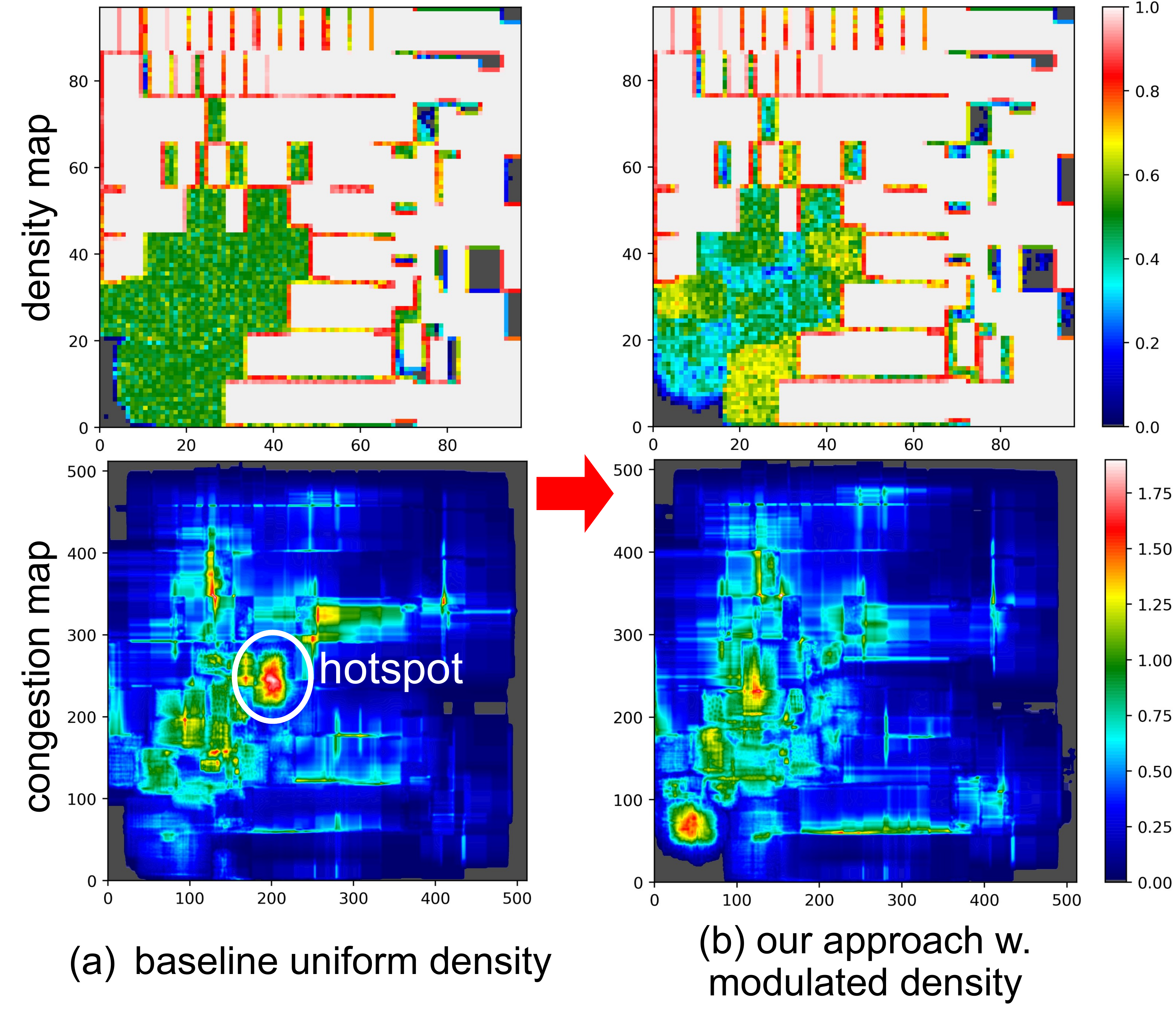}
\caption{Cell density map and overlaid horiz./vert. congestion maps on the Ariane NanGate45 design. The density modulation through cell inflation eliminates the hotspot, resulting in enhanced congestion metrics: (a) peak-weighted-congestion (PWC)=0.85, max/tot. overflow=0.07/0.34 (\%); (b) PWC=0.77, max/tot. overflow 0.04/0.20 (\%).}
\Description[]{}
\label{fig:congestion}
\end{figure}

The DREAMPlace placer that uses ePlace density~\cite{lu2015eplace} can only achieve a uniform density at the end of the global placement. However, congestion is often non-uniform, particularly when the design includes significant routing obstacles like macros, and uniform spreading may result in congestion issues~\cite{alpert2010makes}. Figure~\ref{fig:congestion} illustrates how non-uniform cell placement density can alleviate congestion. Both placements were generated using DREAMPlace with the same parameters, except for our proposed inflation.

The ePlace density loss makes use of the electrostatics Poisson equation to spread the cells
\begin{equation}\label{eq:poisson}
\nabla^2\varphi=-\rho=-\nabla\mathcal{E},
\end{equation}
where $\varphi$ is the electric potential, $\rho$ the cell density, and $\mathcal{E}$ the electric field (=cell spreading force). The solution to \eqref{eq:poisson} results in a null field when the floorplan density is uniform, i.e., $\rho=d_t$, where $d_t$ is the user-defined global target density that dictates the amount of fillers.

While this formulation yields good results for global placement based on wirelength and is especially suitable for numerical computation, it requires modifications to achieve non-uniform density.

\subsubsection{Cell Density Modulation through Inflation}

Larger cells have a greater impact on the density map and spreading field. We increase a cell's size with inflation factor $r_i\geq 1$ to achieve a lower local density than the target density. We only inflate standard cells in the x-direction for legalization purposes, updating pin offsets, and resulting in a new cell width $w_i'=r_i w_i$. We investigated alternative methods for local cell density control, like additional cell-based and bin-based penalties, but found them ineffective due to conflicts with the global Poisson force. In contrast, inflation maintains the original optimization problem, ensuring stable convergence without impacting runtime compared to a wirelength-driven placer. This is critical for successful integration into design space exploration methodologies like AutoDMP.

\noindent\textbf{Inflation Factor Derivation.}
We now present a mathematical derivation of the ``best'' inflation factor. The uniform density leads to $\rho_b\approx d_t, \forall b\in B$. Using \eqref{eq:bin-density} and assuming a standard cell belongs to one bin only and ignoring bins covered by macros, we get
\begin{equation}\label{eq:area-balance}
d_t\approx\rho_b=\frac{1}{A_b}\left(\sum_{i\subset b}r_i a_i + A_{\textrm{fillers}\subset b}\right),
\end{equation}
where the area of fillers inside the bin is set aside. However, we are interested in the effective bin density without inflation, namely
\begin{equation}\label{eq:real-density}
\tilde{\rho}_b=\sum_{i\subset b} \frac{a_i}{A_b}.
\end{equation}
Moreover, combining \eqref{eq:area-balance} and \eqref{eq:real-density} leads to the bounds
\begin{equation}\label{eq:bounds}
\frac{d_t-A_{\mathrm{fillers}\subset b}/A_b}{\max_{i\subset b}\{r_i\}} \le \tilde{\rho}_b \le \frac{d_t-A_{\mathrm{fillers}\subset b}/A_b}{\min_{i\subset b}\{r_i\}}.
\end{equation}

This formula remains true regardless of the choice of inflation. However, to meet the density target for each cell $i$, denoted by $t_i$ (=any of $z_i,\hat{\mu}_i^{(\mathrm{JS})},\hat{\mu}_i^{(\mathrm{JS/D})}$), one must satisfy $\rho_i=\tilde{\rho}_{b|i\subset b} \approx t_i, \forall i\in V$. We must set $d_t=1$ to achieve all target densities in the range $]0,1]$. In a dense placement region devoid of fillers or macros---most important case where cells are tightly pulled together but should be spread out---meeting the targets requires satisfying the simplified \eqref{eq:bounds}
\begin{equation}
1\big/\max_{i\subset b}\{r_i\}\leq t_i\leq 1\big/\min_{i\subset b}\{r_i\}, \, \forall i\subset b,
\end{equation}
which highlights a natural candidate for the inflation factor
\begin{equation}\label{eq:factor}
r_i=1\big/t_i.
\end{equation}

\noindent\textbf{Target Matching Error.} 
Using \eqref{eq:factor} as our inflation factor gives
\begin{equation}
\min_{i\subset b}\{t_i\}\leq\tilde{\rho}_b\leq\max_{i\subset b}\{t_i\},
\end{equation}
which means cells in the same bin must have similar targets to ensure that all target cell densities are met. Thus, note that our inflation steers ePlace's density optimization towards the target without guaranteeing precise mathematical achievement. Although enforcing a strict zero per-bin target range is unfeasible with inflation, this is an acceptable trade-off because it allows the placer to optimize wirelength concurrently with density. We define the per-bin target range
\begin{equation}\label{eq:range}
\max_{i\subset b}\{t_i\}-\min_{i\subset b}\{t_i\}.
\end{equation}
Wide ranges might lead to undesirable outcomes: cells with high-density targets reach lower density when they should be packed to reduce wirelength, whereas low-density cells reach higher density when they should be spread out to improve congestion and leave space for timing optimization. We also define the total range error for later evaluation
\begin{equation}\label{eq:range-error}
\textrm{err.} = \sum_{b \in B} \Big(\max_{i\subset b}\{t_i\}-\min_{i\subset b}\{t_i\}\Big)\quad\textrm{for}\,\, b|\operatorname{Avg}\{t_i\}_{i\subset b} \leq \tilde{\rho}_b,
\end{equation}
which considers only the bins violating the average target cell density.

\subsection{AutoDMP Integration}
AutoDMP adds to DREAMPlace enhanced concurrent macro and standard cell placement and automatic parameter tuning based on multi-objective Bayesian optimization considering wirelength, uniform cell density, and congestion trade-offs. To find high-quality placements with improved congestion and timing, we present how we integrate GOALPlace inside AutoDMP.

\subsubsection{Hellinger Distance}
We introduce an additional objective in AutoDMP: find placements that closely align with the target cell density distribution. For the multi-objective Bayesian optimization, our new axis compares the achieved cell density distribution of the placement $D_{P}$ with the target distribution $D_{T}$ (=histogram of $\boldsymbol{z}$,$\,\boldsymbol{\hat\mu^{(\mathrm{JS})}}$, or $\boldsymbol{\hat\mu^{(\mathrm{JS/D})}}$). We use the symmetric Hellinger distance~\cite{beran1977minimum} computed on the 100-bin histograms of the cell densities
\begin{equation}\label{eq:hellinger}
H(D_{P},D_{T})=\frac{1}{\sqrt{2}}\Big\Vert\sqrt{D_{P}}-\sqrt{D_{T}}\Big\Vert_2\in[0, 1],
\end{equation}
which we wish to minimize. Note that this loss only guides the search, and many different-looking placements exhibit similar density distributions---an important observation of Section~\ref{subsec:cluster}.

\subsubsection{Density Distribution Shift}
To explore the trade-off between congestion, density, and wirelength, rather than always trying to match the target density distribution exactly, we shift the target cell densities using an additional tuning parameter searched in the range $[-0.2,0.2]$. This parameter serves as another objective to minimize in AutoDMP, essentially replacing the uniform density target $d_t$, which is now fixed at 1 for cell inflation. We also use the newly shifted target distribution in the Hellinger distance.

\section{Experiments}\label{sec4}

In our experiments, we compare our methodology with the default AutoDMP across multiple scenarios: industrial and academic (TILOS~\cite{cheng2023assessment}) benchmarks and PnR flows; technologies (commercial 5nm and academic NanGate 45nm/ASAP 7nm PDKs); fixed and unfixed macros; and feeding the macro and/or standard cell placement to a commercial EDA tool for downstream steps. AutoDMP was run on an NVIDIA DGX system with four A100 GPUs, each with 80GB memory, collecting 1,000 placement candidates per benchmark and method. The search runtime of AutoDMP is unaffected by the inflation. We use Cadence Innovus as the commercial tool.

\subsection{Validation of Assumptions}\label{subsec:verif}

\begin{figure}[t]
\centering 
\includegraphics[width=\columnwidth]{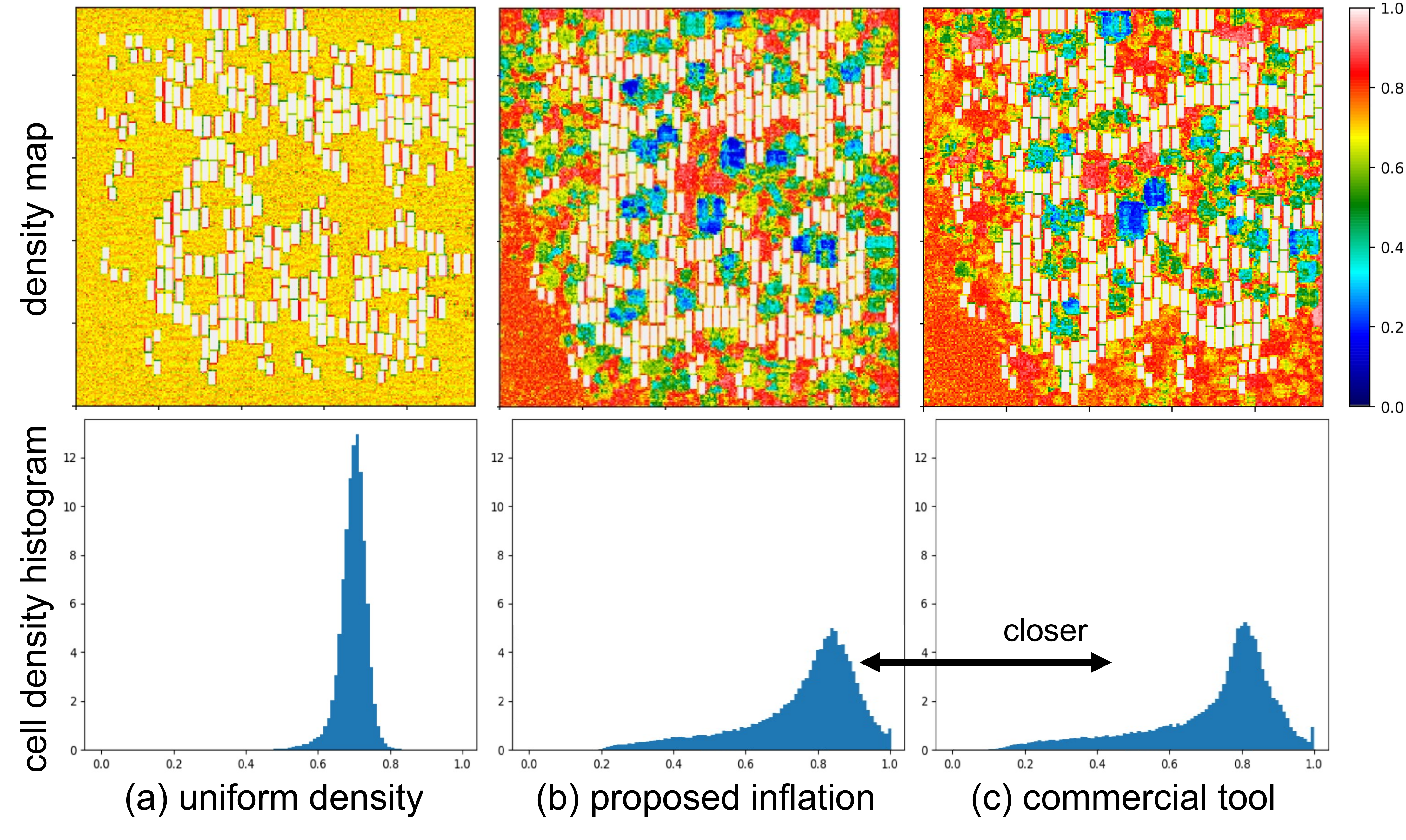}
\caption{Matching a commercial EDA tool's post-route optimized cell density with the proposed cell inflation on the MemPool Group NanGate45 benchmark with movable macros.}
\Description[]{}
\label{fig:mempool}
\end{figure}

\begin{figure}[t]
\centering
\includegraphics[width=.87\columnwidth]{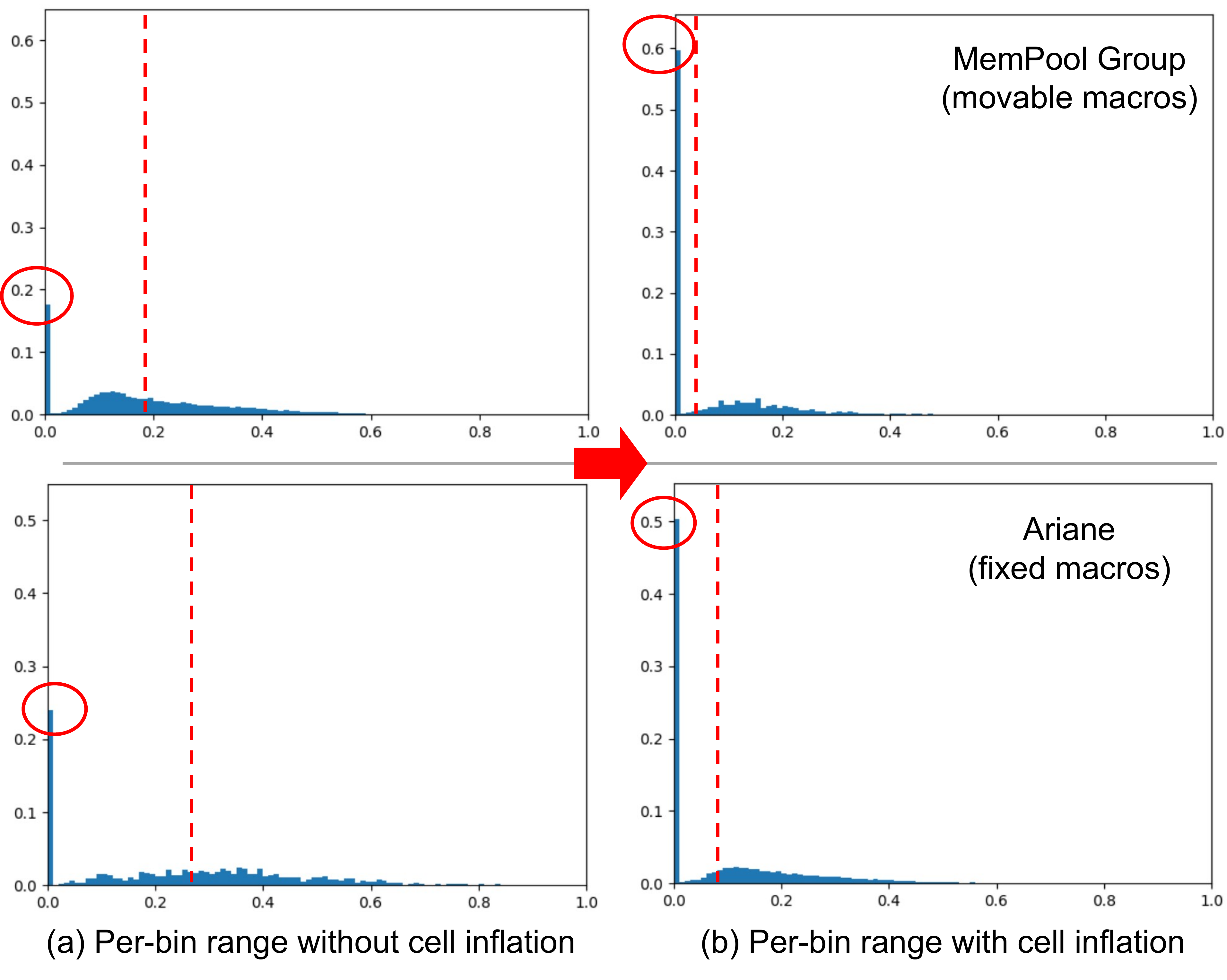}
\caption{Per-bin target cell density ranges \eqref{eq:range} of bins violating the average target density. The red dotted line (=mean error \eqref{eq:range-error}) indicates that the proposed inflation notably narrows these ranges, effectively controlling local cell density.}
\Description[]{}
\label{fig:ranges}
\end{figure}

\begin{table}[t]
\centering\small
\setlength{\tabcolsep}{3pt}
\caption{Correlation analysis between target and achieved densities at cell and cluster levels for the top 50 AutoDMP Pareto placements, without and with inflation using the original tool's target $\boldsymbol{z}$ or the James--Stein (JS) target $\boldsymbol{\hat\mu^{(\mathrm{JS})}}$. Our inflation drives a high correlation, with large improvements from James--Stein (average of 22\% at cell-level and 10\% at cluster-level). Moreover, the high correlation at the cluster level confirms DREAMPlace's clustering effect.}\label{tab:cluster-check}
\begin{adjustbox}{width=\columnwidth}
\begin{tabular}{|c||c|c|c|c|c|}
\hline
\rowcolor{NVfluorite}
&&\multicolumn{2}{c|}{\textbf{Infl. Tool ($\boldsymbol{z}$)}}&\multicolumn{2}{c|}{\textbf{Infl. JS ($\boldsymbol{\hat\mu^{(\mathrm{JS})}}$)}}\\
\rowcolor{NVfluorite}\multirow{-2}{*}{\textbf{Design}}&\multirow{-2}{*}{\textbf{No Infl. (uniform)}}&\textbf{cell}&\textbf{cluster}&\textbf{cell}&\textbf{cluster} \\
\hline\hline
Ariane&$<0$&0.65&0.90&0.72&0.95\\
NVDLA&$<0$&0.61&0.85&0.89&0.97\\
BlackParrot&$<0$&0.58&0.83&0.71&0.92\\
MemPool Group&$<0$&0.76&0.92&0.85&0.99\\
\hline
\end{tabular}
\end{adjustbox}
\end{table}

Figure~\ref{fig:mempool} visually demonstrates the effectiveness of cell density control using our proposed cell inflation. Despite variations in macro placement, DREAMPlace's macro and standard cell placements closely align with the cell density distribution and appearance of the density map from the commercial EDA tool used to generate the targets.

\subsubsection{Bin Level Evaluation}
To validate the algorithmic success of the inflation, we evaluate the size of the total range error \eqref{eq:range-error} at the bin-level. Figure~\ref{fig:ranges} shows the significant reduction in the ranges of each bin across placements and designs with the implementation of cell inflation, indicating most target densities are met on average. Specifically, the total range error is reduced significantly by 10$\times$ on MemPool and 4$\times$ on Ariane. Using the James--Stein adapted target densities leads to further reductions in the ranges (not depicted).

\subsubsection{Cell/Cluster Level Evaluation}
Table~\ref{tab:cluster-check} shows the correlation between the target and achieved densities in the top 50 AutoDMP Pareto placements, revealing a strong cell-level correlation that underscores the effectiveness of the proposed inflation, aligning with the bin-level evaluations. Employing the James--Stein adapted targets \eqref{eq:JS} instead of the original tool targets enhances cell-level correlation by an average of 22\%, indicating better alignment with DREAMPlace outputs. Notably, correlations at the cluster level are incredibly high, supporting the hypothesis of a clustering effect in DREAMPlace. This suggests that the proposed cell-level inflation method also effectively manages density more globally as mentioned in Section~\ref{subsec:cluster}, aiding in scenarios requiring a more global spreading.

\subsection{Global Placement Quality Results} 

\begin{table}[t]
\centering\small
\setlength{\tabcolsep}{3pt}
\caption{Post-place average-quality comparison of top 50 Pareto macro and standard cell placements candidates from AutoDMP variants. The average wirelength, congestion, and density hotspots are reported using the commercial tool's early global router. The last column reports the number of top 5 Pareto points from each method, recalculated based on the more accurate tool's metrics.}\label{tab:compare-paretos}
\begin{adjustbox}{width=\columnwidth}
\begin{tabular}{|c||c|c|c|c|c|}
\hline
\rowcolor{NVfluorite}
&\textbf{WL}&\textbf{HPWL}&\textbf{Cong. H/V}&\textbf{Density}&\textbf{Updated}\\
\rowcolor{NVfluorite}
\multirow{-2}{*}{\textbf{Method}}&\textbf{(m)}&\textbf{(m)}&\textbf{(\%)}&\textbf{Hotspot}&\textbf{Paretos}\\
\hline\hline
\multicolumn{6}{|c|}{\textit{\textbf{NVDLA-NG45} (150K cells, 128 macros, 1.11 GHz, \textbf{movable} macros)}} \\ 
\hline
AutoDMP &9.25&6.69&0.09/0.39&44.5&0/5\\
AutoDMP w/ pin infl.&9.18&6.65&0.08/0.41&46.1&0/5\\
Ours&\underline{\textbf{9.09}}&6.58&0.07/0.31&44.6&\ubf{2/5}\\
Ours w/ James--Stein&\ubf{9.08}&\ubf{6.54}&\ubf{0.05/0.28}&\ubf{43.2}&\ubf{3/5}\\
\hline
\multicolumn{6}{|c|}{\textit{\textbf{BlackParrot-NG45} (650K cells, 220 macros, 769 MHz, \textbf{movable} macros)}}\\ \hline
AutoDMP &25.21&19.63&0.08/0.24&57.3&1/5\\
AutoDMP w/ pin infl.&24.92&19.41&0.08/0.26&62.1&0/5\\
Ours&24.26&\ubf{18.94}&0.06/0.19&55.0&1/5\\
Ours w/ James--Stein&\ubf{24.20}&19.12&\ubf{0.06/0.16}&\ubf{53.5}&\ubf{3/5} \\
\hline
\end{tabular}
\end{adjustbox}
\end{table}

We evaluate the global place quality of the top 50 Paretos points from default AutoDMP and AutoDMP with GOALPlace on NVDLA and BlackParrot designs with movable macros. We also compare them with the default AutoDMP flow augmented with an extra tunable parameter ($\alpha$) for uniform pin density inflation, whereby each cell is inflated to maintain the same number of sites per pin, following
\begin{equation}
w_i'=\left[\operatorname{sites}(i)+\max\left(\left\lceil\alpha\frac{\#\operatorname{pins}(i)}{\# \operatorname{rows}(i)}-\operatorname{sites}(i)\right\rceil,0\right)\right] w_{\operatorname{site}}.
\end{equation}
We also attempted to use the routability-driven RePlace flow~\cite{cheng2018replace} from DREAMPlace but faced challenges in achieving satisfactory convergence on benchmarks with movable macros. As noted in Section~\ref{subsec:cong-est}, these iterative methods relying on congestion estimation are slow and tricky to tune. The commercial tool is also expected to implement similar techniques of much higher quality.

Table~\ref{tab:compare-paretos} evaluates macro and standard cell placements for density hotspots, wirelength, and congestion using a commercial tool's early global router. \textit{Ours} refers to GOALPlace with the original tool's target and \textit{Ours w/ James--Stein} to GOALPlace with the James--Stein target. The ``Updated Paretos'' column shows the top 5 candidates recalculated using the tool's metrics. Our methodology significantly improves the overall quality of candidates. In contrast, due to its lack of placement awareness, the pin inflation method results in much higher density and congestion. Conversely, the James--Stein adaptation markedly improves the wirelength-density-congestion trade-off, reducing the average Hellinger distance \eqref{eq:hellinger} from 0.25 to 0.11.

\subsection{Post-Route Quality Results} 

\begin{table*}[t]
\centering\small
\setlength{\tabcolsep}{3pt}
\caption{Post-route PPA comparison between default AutoDMP and GOALPlace on the TILOS flow and benchmark~\cite{cheng2023assessment}. \textit{Macro}/\textit{Std-Cell} indicates if macro and/or standard cell placements are used for downstream steps in the commercial tool. On MemPool, the best default AutoDMP parameters were augmented with cell inflation without repeating the search (*). The James--Stein experiments require two AutoDMP runs, including one using the tool's target. These results highlight the effectiveness of our inflation with and without James--Stein adaptation across various scenarios. We highlight the metrics that are significantly better than the commercial tool or baseline default AutoDMP.}\label{tab:results-autodmp}
\begin{adjustbox}{width=\textwidth}
\begin{tabular}{|c||c|c|c|c|c|c|Hc|c|c|c|}
\hline
\rowcolor{NVfluorite}
&\textbf{WL}&\textbf{Power}&\textbf{WNS all}&\textbf{TNS all}&\textbf{WNS r2r}&\textbf{TNS r2r}&\textbf{Cong. H/V}&\textbf{Cong. H/V}&\textbf{\#}&\textbf{Search}&\textbf{Tool}\\
\rowcolor{NVfluorite}
\multirow{-2}{*}{\textbf{Method}}&\textbf{(m)}&\textbf{(mW)}&\textbf{(ns)}&\textbf{(ns)}&\textbf{(ns)}&\textbf{(ns)}&\textbf{Pre-CTS (\%)}&\textbf{Pre-Route (\%)}&\textbf{DRCs}&\textbf{Runtime}&\textbf{Runtime}\\
\hline\hline
\multicolumn{12}{|c|}{\textit{\textbf{MemPool Group-NG45} (2.7M cells, 320 macros, 333 MHz, \textbf{movable} macros)}} \\ 
\hline
Tool&115.5&4124&-0.398&-2668&-0.206&-1548&3.59/1.88&3.52/1.76&2132&-&32h\\
\hdashline
AutoDMP (\textit{Macro})&111.1&4091&-0.330&-2913&-0.226&-1955&3.48/1.86&3.32/1.66&2651&3h30&28h\\
Ours (\textit{Macro})&\ubf{110.2}&\ubf{4066}&-0.318&-2899&-0.178&-1892&\ubf{2.56/1.73}&\ubf{2.44/1.54}&\ubf{210}&$*$&24h\\
Ours w/ James--Stein (\textit{Macro})&\ubf{109.8}&\ubf{4071}&-0.410&-3688&-0.207&-2711&&\ubf{2.22/1.57}&\ubf{144}&2$\times$3h&26h\\
Ours w/ James--Stein w/ Timing (\textit{Macro})&\ubf{110.4}&\ubf{4064}&\ubf{-0.303}&\ubf{-2434}&\ubf{-0.130}&\ubf{-1339}&&\ubf{2.31/1.59}&\ubf{142}&2$\times$3h&24h\\
\hdashline
AutoDMP (\textit{Macro}/\textit{Std-Cell})&112.0&4087&-0.463&-4708&-0.196&-3438&3.24/2.30&2.87/1.87&8189&3h30&33h\\
Ours (\textit{Macro}/\textit{Std-Cell})&111.9&4075&-0.355&-4072&-0.169&-3117&\ubf{2.27/1.97}&\ubf{2.00/1.59}&\ubf{59}&*&27h\\
Ours w/ James--Stein (\textit{Macro}/\textit{Std-Cell})&\ubf{110.7}&\ubf{4056}&-0.398&-4164&-0.198&-3397
&&\ubf{2.12/1.53}&\ubf{76}&2$\times$3h&27h\\
Ours w/ James--Stein w/ Timing (\textit{Macro}/\textit{Std-Cell})&\ubf{110.7}&\ubf{4051}&\ubf{-0.327}&\ubf{-3586}&\ubf{-0.116}&\ubf{-1390}&&\ubf{2.24/1.64}&\ubf{60}&2$\times$3h&25h\\
\hline
\multicolumn{12}{|c|}{\textit{\textbf{Ariane-NG45} (100K cells, 133 macros, 769 MHz, \textbf{fixed} tool's macro placement)}} \\ 
\hline
Tool&4.120&825&-0.183&-134&-0.183&-134&0.02/0.02&0.02/0.02&0&-&1h\\
AutoDMP (\textit{Std-Cell})&4.056&825&-0.218&-185&-0.218&-185&0.03/0.13&0.05/0.15&0&50m&1h10\\
Ours (\textit{Std-Cell})&4.028&825&-0.204&-195&-0.204&-195&0.02/\textbf{0.03}&0.03/0.08&0&35m&1h\\
Ours w/ James--Stein (\textit{Std-Cell})&\ubf{3.989}&822&-0.210&-188&-0.210&-188&&0.02/0.04&0&2$\times$35m&1h\\
Ours w/ James--Stein w/ Timing (\textit{Std-Cell})&\ubf{4.001}&823&\ubf{-0.168}&\ubf{-120}&\ubf{-0.168}&\ubf{-120}&&0.03/0.05&0&2$\times$35m&1h\\
\hline
\multicolumn{12}{|c|}{\textit{\textbf{BlackParrot-NG45} (650K cells, 220 macros, 769 MHz, \textbf{fixed} DREAMPlace's macro placement)}}\\ 
\hline
Tool&24.55&4464&-0.287&-4992&-0.287&-4992&0.04/0.11&0.05/0.13&0&-&5h40\\
AutoDMP (\textit{Std-Cell})&24.66&4483&-0.281&-2180&-0.281&-2180&0.04/0.15&0.05/0.17&0&1h50&5h20\\ 
Ours (\textit{Std-Cell})&24.42&4468&-0.226&-2949&-0.226&-2949&0.03/0.09&0.03/0.10&0&1h30&4h50\\
Ours w/ James--Stein (\textit{Std-Cell})& \ubf{24.28}&4471&-0.236&-2205&-0.236&-2205&0.03/0.07&\ubf{0.02/0.08}&0&2$\times$1h30&5h\\
Ours w/ James--Stein w/ Timing (\textit{Std-Cell})&\ubf{24.23}&4466&\ubf{-0.198}&\ubf{-1974}&\ubf{-0.198}&\ubf{-1974}&\ubf{0.02/0.00}&\ubf{0.02/0.08}&0&2$\times$1h30&5h20\\
\hline
\multicolumn{12}{|c|}{\textit{\textbf{Ariane-AS7} (100K cells, 133 macros, 1.11 GHz, \textbf{movable} macros)}}\\ 
\hline
Tool&0.921&503&-0.024&-0.752&-0.024&-0.752&0.08/0.05&0.18/0.04&3&-&1h\\
AutoDMP (\textit{Macro})&0.928&507&-0.045&-4.474&-0.045&-4.474&0.06/0.04&0.11/0.04&166&1h&1h40\\ 
Ours (\textit{Macro})&0.926&504&-0.035&-0.637&-0.035&-0.637&0.08/0.01&0.09/0.04&60&50m&1h30\\ 
Ours w/ James--Stein (\textit{Macro})&\ubf{0.894}&503&-0.017&-0.799&-0.017&-0.799&0.06/0.02&\ubf{0.04/0.01}&19&2$\times$50m&1h10\\
Ours w/ James--Stein w/ Timing (\textit{Macro})&\ubf{0.901}&503&-0.020&\ubf{-0.211}&-0.020&\ubf{-0.211}&0.06/0.03&\ubf{0.05/0.02}&32&2$\times$50m&1h\\
\hline
\end{tabular}
\end{adjustbox}
\end{table*}

Table~\ref{tab:results-autodmp} shows a post-route power-performance-area (PPA) comparison between the default AutoDMP and our methodology on the TILOS flow and benchmarks. As in the original AutoDMP methodology, we only report the best PPA among the top five Pareto candidates pushed to post-route. GOALPlace achieves quality superior or comparable to commercial tools, significantly improving PPA over AutoDMP in various scenarios, including the difficult movable macro scenario and across a spectrum of less to more congested designs. Additionally, the James--Stein adaptation yields further wirelength benefits and notable timing improvements with our timing enhancement $\boldsymbol{\hat\mu^{(\mathrm{JS/D})}}$ (mention \textit{Ours w/ James--Stein w/ Timing}). The detailed observations are:
\begin{itemize}
\item AutoDMP fails to produce routable placements on the very congested MemPool Group design. Contrarily, our methodology produces macro and standard cell placements (Figure~\ref{fig:mempool}) with significant congestion reduction, leading to more than 10$\times$ fewer post-route DRC violations. Note that for MemPool, in one scenario, the cell inflation was added to the best default AutoDMP parameters without rerunning a search, highlighting the limited tuning needed for our method's success.
\item The results for Ariane and BlackParrot NanGate45 designs with fixed macro placement obtained from different sources (the tool and DREAMPlace) show that our methodology consistently delivers high-quality standard cell placements with substantial reductions in wirelength and congestion compared to default AutoDMP, regardless of how the macro placement was generated.
\item The results of the Ariane implementation on ASAP 7nm show the applicability of our method across technologies and notable wirelength improvements from the James--Stein adaptation.
\item The inflation with James--Stein target consistently improves congestion and wirelength relative to the baseline, though it may sometimes worsen TNS. As noted in Section~\ref{subsec:timing}, timing-critical cells can see excessive target density increases under rule \eqref{eq:JS}. However, the enhancement of \eqref{eq:translation} compensates for this degradation, providing large WNS and TNS gains of up to 30\% and 60\%.
\end{itemize}

\begin{table}[t]
\centering\small
\setlength{\tabcolsep}{3pt}
\caption{Post-route average metrics of top 10 Pareto macro placements on a densely packed industrial design using a 5nm PnR flow. Results are timing and DRC clean, assessed using an internal evaluation tool.}\label{tab:industrial-bench}
\begin{adjustbox}{width=\columnwidth}
\begin{tabular}{|c||c|c|c|c|}
\hline
\rowcolor{NVfluorite}
\textbf{Method}&\textbf{WL}&\textbf{Best WL}&\textbf{Max-Cong.}&\textbf{Best Max-Cong.}\\
\hline\hline
\multicolumn{5}{|c|}{\textit{\textbf{ML Accelerator} (400K cells, 65\% util., 48 \textbf{movable} macros)}}\\ 
\hline
Tool&-&2.11&-&0.58\\
AutoDMP&2.27&2.14&0.63&0.61\\
Ours&2.15&2.11&0.56&0.54\\
Ours w/ James--Stein&2.13&\ubf{2.09}&\ubf{0.55}&\ubf{0.51}\\
\hline
\end{tabular}
\end{adjustbox}
\end{table}

Table~\ref{tab:industrial-bench} illustrates that our methodology effectively handles a 5nm industrial design with complex detailed rules, achieving a 5\% reduction in post-route wirelength compared to default AutoDMP. This improvement is mainly due to significant improvements in maximum congestion, averaging 11.1\% across the top 10 Pareto macro placements. Since macros comprise 45\% of the floorplan utilization in this highly congested design, spreading them without creating congestion hotspots is challenging. This emphasizes the importance of our placement/congestion co-optimization method in advanced technologies, where managing congestion is crucial for wirelength.

\section{Conclusion}\label{sec5}
We propose a new learning-based approach based on cell density control to improve placement quality. The importance of density is confirmed by a statistical analysis based on a new hierarchical netlist clustering, highlighting the strong correlation between placement, density, and timing. Our methodology includes new critical ideas. First, we propose to learn from an EDA tool's post-route optimized cell densities to capitalize on their advanced routability and timing optimization. Second, by taking an empirical Bayes approach, we show how to combine Bayesian and frequentist ideas for large-scale statistical inference to adapt the density targets to the specific placer's densities. Integrated inside an academic GPU-accelerated global placer and Bayesian search, our method delivers placements comparable to commercial tools across all PPA metrics. We hope this research underscores the potential of learning from previous PnR runs' final results and the practicality of statistical inference, inherited from Bayes's thinking, as a catalyst for future design flows. In the future, we aim to accelerate GOALPlace by learning a model to predict the post-route optimized or adapted cell-level density without running the EDA tool. This could be especially useful when working with small amounts of data during the early design exploration phase, where the netlist changes significantly. Last, we intend to explore how to incorporate our approach into a positive feedback loop with EDA tools, where the density target is progressively improved based on the changed density distribution of their output.

% \clearpage
% \balance
\bibliographystyle{ACM-Reference-Format}
\bibliography{biblio}

\end{document}